\documentclass[acmtog]{acmart} 

\usepackage[table]{xcolor}
\usepackage{booktabs} 
\usepackage{makecell}
\usepackage{balance}

\usepackage{graphicx,enumitem}
\usepackage{stfloats}
\usepackage{multirow}

\usepackage[ruled]{algorithm2e} 

\SetAlFnt{\small}
\SetAlCapFnt{\small}
\SetAlCapNameFnt{\small}
\SetAlCapHSkip{0pt}
                                
\acmJournal{TOG}
\acmYear{2025}
\acmVolume{44}
\acmNumber{6}
\acmArticle{263}
\acmMonth{12}
\acmDOI{10.1145/3763303}

\setcopyright{none}
\renewcommand\footnotetextcopyrightpermission[1]{}
\makeatletter
\let\@ACM@journal@footnotes\@empty
\makeatother

\begin{document}
\title{ConsiStyle: Style Diversity in Training-Free Consistent T2I Generation}

\begin{teaserfigure}
  \centering
  \includegraphics[width=0.91\textwidth]{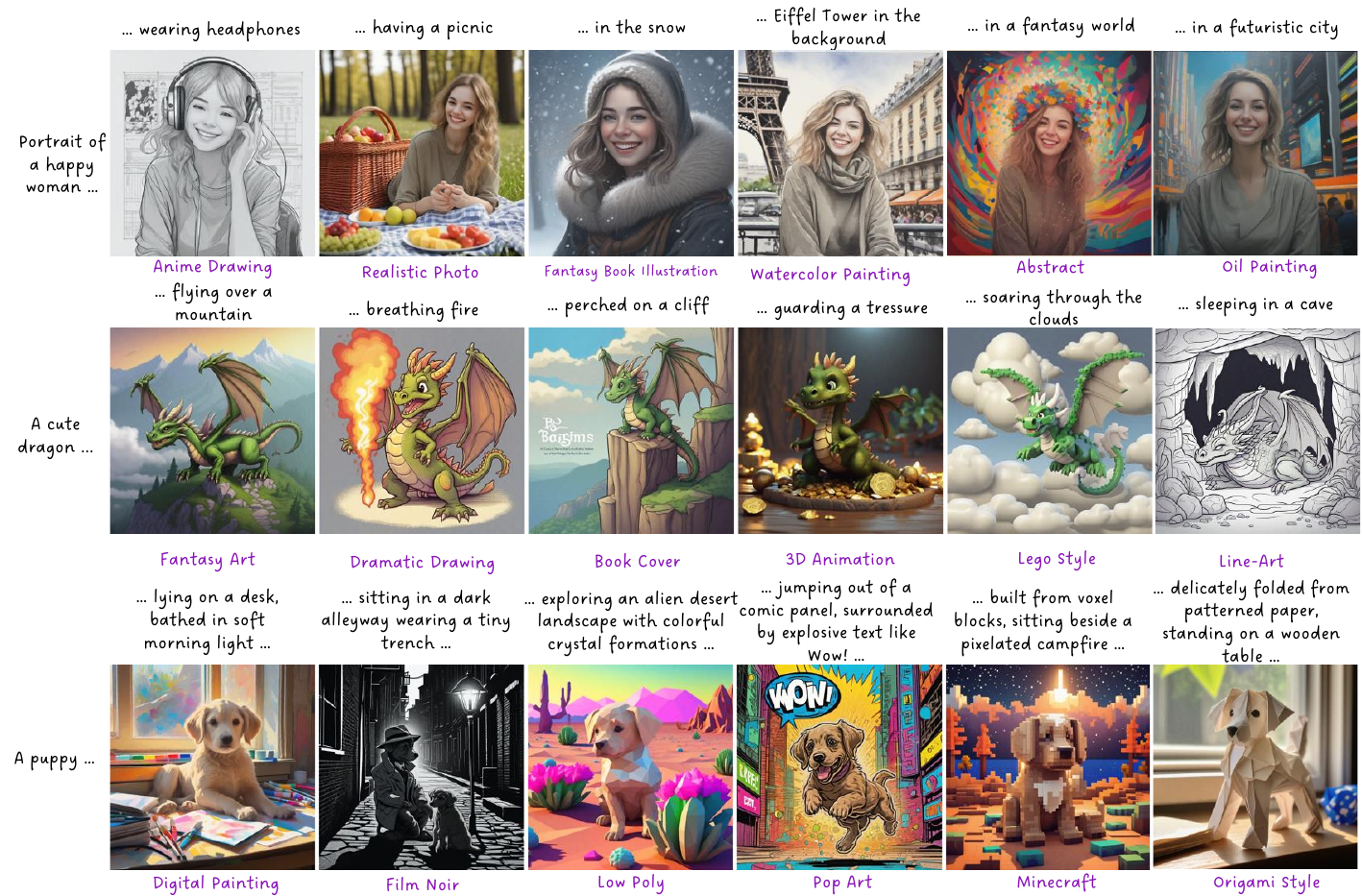}
  \caption{
    We present \textbf{ConsiStyle} - a training-free optimization method that decouples style from subject-specific characteristics such as color, structure, patterns and unique markings. Our approach preserves subject consistency across various prompts while maintaining alignment with diverse style descriptions.
  \label{fig:teaser}}
\end{teaserfigure}

\author{Yohai Mazuz}
\email{yohaimazuz@mail.tau.ac.il}  
\authornote{Denotes equal contribution.}
\orcid{0009-0004-0574-6546}
\affiliation{%
 \institution{Tel Aviv University}
 \country{Israel}
 }

\author{Janna Bruner}
\email{jannabruner@mail.tau.ac.il}
\authornotemark[1]
\orcid{1234-5678-9012-3456}
\affiliation{%
 \institution{Tel Aviv University}
 \country{Israel}
}
 
\author{Lior Wolf}
\email{wolf@cs.tau.ac.il}
\orcid{0009-0004-8582-6182}
\affiliation{%
 \institution{Tel Aviv University}
 \country{Israel}
 }



\begin{abstract}

In text-to-image models, consistent character generation is the task of achieving text alignment while maintaining the subject's appearance across different prompts. However, since style and appearance are often entangled, the existing methods struggle to preserve consistent subject characteristics while adhering to varying style prompts. Current approaches for consistent text-to-image generation typically rely on large-scale fine-tuning on curated image sets or per-subject optimization, which either fail to generalize across prompts or do not align well with textual descriptions. Meanwhile, training-free methods often fail to maintain subject consistency across different styles. 
In this work, we introduce a training-free method that, for the first time, jointly achieves style preservation and subject consistency across varied styles. The attention matrices are manipulated such that Queries and Keys are obtained from the anchor image(s) that are used to define the subject, while the Values are imported from a parallel copy that is not subject-anchored. Additionally, cross-image components are added to the self-attention mechanism by expanding the Key and Value matrices. To do without shifting from the target style, we align the statistics of the Value matrices.
As is demonstrated in a comprehensive battery of qualitative and quantitative experiments, our method effectively decouples style from subject appearance and enables faithful generation of text-aligned images with consistent characters across diverse styles.   

Code will be available at our project page: \href{https://jbruner23.github.io/consistyle/}{jbruner23.github.io/consistyle}.
\end{abstract}

\maketitle

\begin{center}
{\small This is the author version of a paper accepted to SIGGRAPH Asia 2025.

The final published version appears in ACM Transactions on Graphics.}
\end{center}

\section{Introduction}
In visual storytelling, from comics to animation to movies, the same character often traverses diverse stylistic worlds. Whether rendered as a hyper-realistic portrait, a minimal sketch, or even a pixel art figure in a parody sequence, the human eye can perceive subjects of different styles as the same character, see Fig. ~\ref{fig:teaser}. However, since style is a crucial part of the overall appearance, maintaining identity while varying style poses a tremendous technical challenge.

Text-to-image diffusion models~\cite{Rombach2021HighResolutionIS, Saharia2022PhotorealisticTD, Chang2023MuseTG} have made significant progress in generating high-quality, stylized images from text prompts, enabling the creation of diverse and complex visuals. Yet, these models typically generate each image independently, making it difficult to preserve consistent subject identity across multiple images or prompts.

There are three factors we would like to control independently: (i) prompt-aligned scene and setting, (ii) prompt-aligned image style, and (iii) cross-image character consistency. 
These factors have been studied in various partial combinations. 
\citet{Hertz2023StyleAI} have shown that using attention sharing techniques, the style of generated images can be aligned without pre-training, while \citet{alaluf2023crossimage} show how to transfer the appearance of an object in one image to another by mixing attention components between the images. Character consistency has been studied either as a personalization problem \cite{Gal2022AnII, ruiz2022dreambooth} or as a consistent generation problem. The former receives the target subject as a set of input images. The latter only requires that the generated subject is fixed among all generated images, and can be either based on finetuning (reducing the problem to that of personalization)~\cite{avrahami2024chosen, Gong2023InteractiveSV, ryu2023lora} or by training-free approaches~\cite{tewel2024consistory, zhou2024storydiffusion}. Training-free approaches offer prompt-faithful generation, but fall short in maintaining subject consistency across diverse styles.

\begin{figure}[t] %
    \centering
    \includegraphics[width=0.48\textwidth]{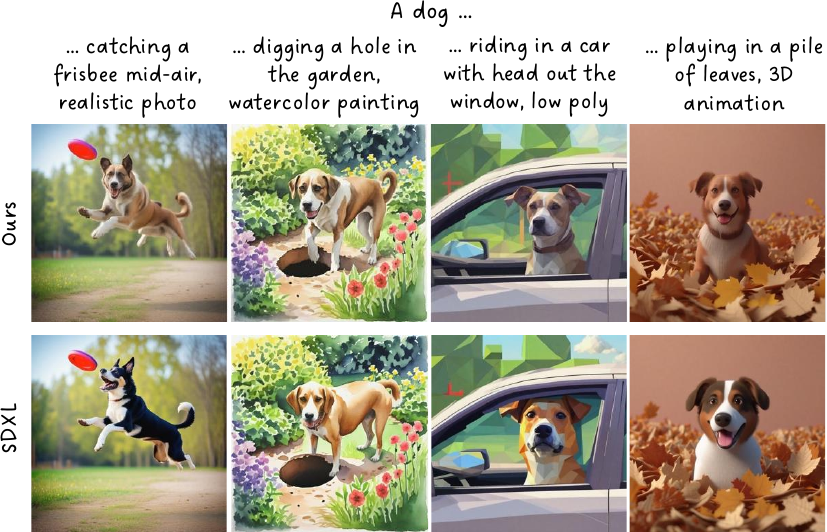}
    \caption{Consistent character generation across diverse styles. Our method preserves key characteristics such as patterns and colors while adhering to the style specified in each prompt. In contrast, SDXL aligns with the prompt and style but fails to maintain consistency across different prompts.}
    \Description{A grid of images comparing results from our method and SDXL. The images show the same character rendered in different styles, highlighting consistency in our results and inconsistency in SDXL.}
    
    \label{fig:intro_figure}
\end{figure}

In this work, we address the problem of generating consistent characters across varying prompts and styles, proposing a training-free framework that aligns both semantic identity and visual style, as shown in Fig.~\ref{fig:intro_figure}. Our method consists of three main stages:
\begin{enumerate}[leftmargin=*]
    \item Style extraction: We run SDXL \cite{sdxl} with the desired prompts and record the Value matrices from the self-attention layers. These serve as style anchors in the later diffusion process.
    \item Cross-image attention with style alignment: We modify the self-attention mechanism to allow each image to attend to the others during generation, encouraging subject consistency across the image set, we apply adaptive instance normalization to avoid style leak between images.
    \item Identity alignment: We compute feature correspondences using DIFT~\cite{Tang2023EmergentCF}, and apply the resulting mappings to the Query and Key matrices only. In the early diffusion steps, the previously recorded Values are injected to guide the process toward the desired style distribution.
\end{enumerate}

As far as we can ascertain, this is the first training-free method to jointly decouple style from identity while ensuring both prompt alignment and subject consistency across diverse styles.

Our empirical evaluations demonstrate that our method consistently outperforms prior approaches in both style and prompt alignment, while maintaining subject consistency comparable to existing methods.

\section{Related Work}

\begin{figure*}[t] %
    \centering
    \includegraphics[width=0.7\paperwidth]{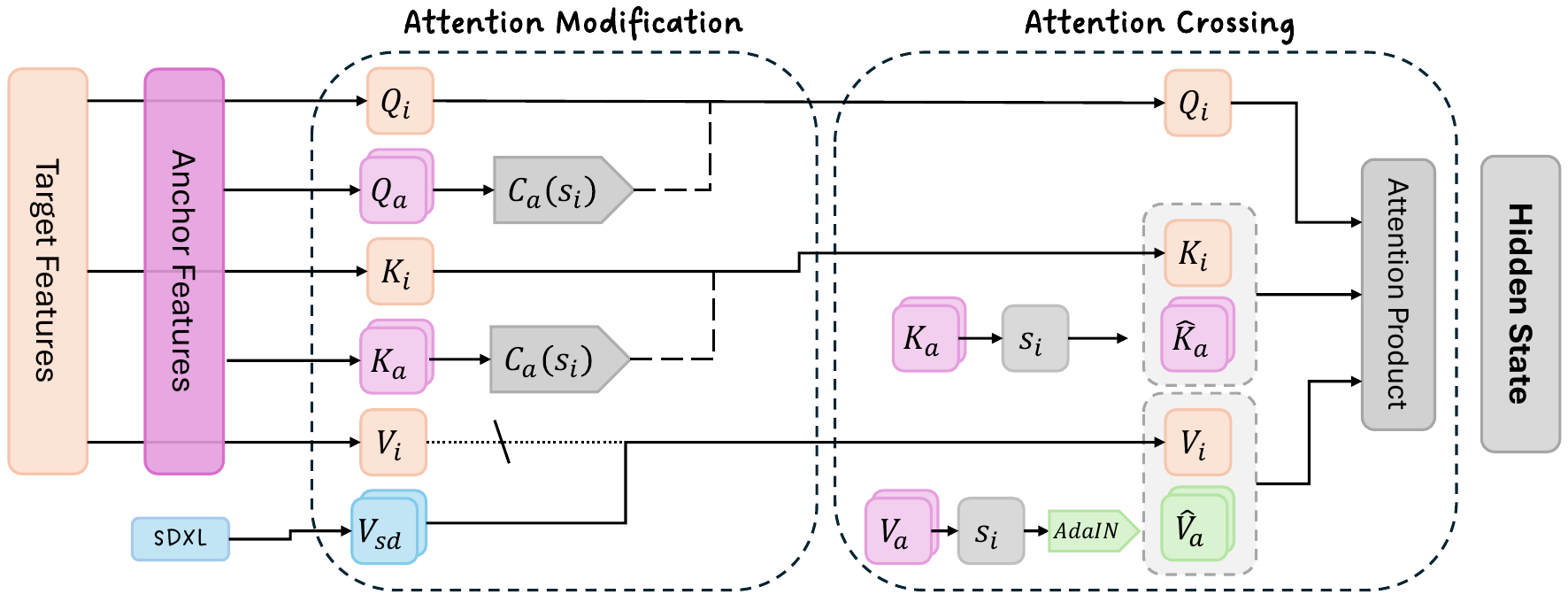}    
    \caption{Overview of our method, illustrating the attention modification and crossing components.}
    \label{fig:method_overview}
\end{figure*}
We aim to generate an array of images depicting consistent subjects across diverse styles, enabling more flexible and expressive prompt design. Following previous work, the main characteristics that are concerned with in style are the shapes, textures and colors.

\textit{Style Transfer and Style Alignment} techniques aim to disentangle visual style from content, enabling the generation of diverse outputs while preserving underlying structure or semantics. Early approaches focused on transferring artistic styles onto photographs \cite{Gatys2016ImageST}, whereas more recent methods have emphasized stronger style-content decoupling, preserving spatial structure and identity across a variety of stylistic domains~\cite{alaluf2023crossimage, Gao2024StyleShotAS, bruner2025illusign}.
In the context of style alignment, Hertz et al.~\cite{Hertz2023StyleAI} propose extending the self-attention mechanism to share attention across a set of images, promoting consistent style across generations. While these methods achieve state-of-the-art results in their respective domains, they are not directly suited to the problem we address. Our work builds upon these advancements, aiming to explicitly distinguish between style and character identity, and to preserve both consistently across varying prompts and visual domains.

\textit{Personalization} methods aim to condition generative models on a specific subject, enabling consistent synthesis of that subject in new contexts. DreamBooth~\cite{ryu2023lora} and Textual Inversion~\cite{Gal2022AnII} introduced approaches to personalize diffusion models using only a few images of a subject. Though effective, these methods require subject-specific training and often struggle with preserving fidelity across diverse prompts or styles.

\textit{Consistent Text-to-Image Generation} maintaining consistency of a character or subject across multiple generated images remains a key challenge in text-to-image diffusion. Some methods leverage attention maps, reference encodings, or optimization strategies to enforce identity coherence across generations~\cite{avrahami2024chosen, tewel2024consistory, zhou2024storydiffusion}. However, many of these approaches either depend on personalization or exhibit limited flexibility when prompts vary significantly in content or style. Ensuring prompt alignment while maintaining consistent identity across diverse visual appearances is still an open problem.

\textit{Image Harmonization} aims to make composite images appear visually coherent by ensuring that the foreground object aligns stylistically with the background scene. This involves matching lighting, texture, and color distributions to produce seamless and natural-looking compositions. Although harmonization primarily focuses on visual realism rather than identity preservation, techniques from this field~\cite{Zhu2022ImageHB, Cong2021DeepIH, Tsai2017DeepIH} inform design choices in style-consistent generation.
In our method, extending the attention mechanism and injecting feature correspondences between foreground subjects can occasionally introduce artifacts, disrupting visual harmony between the foreground and background. To address this, we incorporate the Value matrices extracted from SDXL, this guides the generation process toward the original style and composition distribution, resulting in more naturally harmonized outputs.

\section{Method}

Modern text-to-image (T2I) diffusion models integrate transformer blocks within the U-net layers, allowing patches in the latent space to attend to one another. This process refines image coherence by enabling feature aggregation across the spatial dimensions.

Given input features $X \in \mathbb{R}^{B \times N \times d}$, where $B$ is the batch size, \(N = H \times W\) is the number of patches, and $d$ is the feature dimension, self-attention employs three learnable linear projections to compute the query, key, and value matrices:
\begin{equation}
    Q, K, V \in \mathbb{R}^{B \times N \times d}\,,
\end{equation}
The self-attention mechanism captures the pairwise relationships between patches using the scaled dot-product formulation:
\begin{equation}
\text{Attention}(Q_i, K_i, V_i) = \text{softmax}\left(\frac{Q_i K_i^\top}{\sqrt{d}}\right) V_i\,,
\end{equation}
\noindent where $i \in [B]$ is the index of an image in the batch. This aggregates context from the entire image, enhancing feature representations for more consistent and contextually accurate outputs. The resulting attended features are then typically projected back to the original feature dimension before being passed to subsequent layers.

\subsection{Method Overview}

Maintaining subject consistency in training-free approaches for diffusion models remains a significant challenge. Existing methods often rely on injecting hidden states or cross-image key and value sharing within the self-attention layers to preserve subject identity. However, these approaches can inadvertently introduce style misalignment, as hidden states typically encode both semantic and visual features. For instance, injecting the hidden state of a colorful image into a grayscale context can result in unintended color transfer, leading to stylistic inconsistencies.

To address this, our approach focuses on isolating the semantic consistency of subjects while reducing unintended style entanglement. By precisely managing the flow of visual features and separating semantic content from stylistic elements, our method ensures accurate subject alignment without compromising intended appearance. This balance is accomplished through a combination of targeted attention mechanisms and adaptive normalization, designed to preserve structural integrity while maintaining style fidelity.

Our approach for improving style and subject consistency in text-to-image diffusion models is illustrated in Fig.~\ref{fig:method_overview}. It is a multi-phase process (see Table ~\ref{tab:symbols} for a list of the symbols): 

\begin{enumerate}
    \item \textbf{Initial generation ~~} We first run a vanilla generation using the SDXL model. During this pass, we store the intermediate value features $V_{sd}$, which capture the fine-grained texture and color details needed for consistent style preservation in subsequent generations.

    \item \textbf{Correspondence computation~~} 
Next, we run a generation to compute a set of DIFT features \cite{Tang2023EmergentCF}, which are used to establish a {correspondence mapping} $C_\alpha$ for the subject indices $s_{\alpha}$ in each image $\alpha$ and only includes attention crossing, a component which allows images to attend to subjects of other images in the self-attention layer and does not rely on the correspondence mapping. The subject indices are obtained from the cross-attention layer using a threshold over the attention map matching the subject token query. The images generated for computing DIFT do not employ the $Q$ and $K$ modifications of Sec.~\ref{sec:attn_transfer} since these require the subject location information that is computed using DIFT. However, the component crossing of Sec.~\ref{ref:adain}, in which $K$ and $V$ are enriched with key and value pairs from other images in the batch after applying AdaIn is applied to obtain some level of subject consistency while maintaining style.

    \item \textbf{Final generation~~} 
In the final pass, we perform a full generation that integrates all components: \textbf{i. Value-Preservation:} We reuse the stored values from the initial vanilla run, maintaining stylistic consistency,  \textbf{ii. Attention Transfer:} During the initial phase, we employ the correspondence mapping to inject query and key features, for aligning the subject details across images, and
\textbf{iii. Attention Crossing:} Throughout the final generation, we apply attention crossing to allow image queries to attend to the Keys and Values of other images in the batch, which improves subject consistency, while using AdaIN to prevent style-leak between images due to the different Value distributions.

\end{enumerate}

\begin{figure*}[t] %
    \centering
    \includegraphics[width=0.9\textwidth]{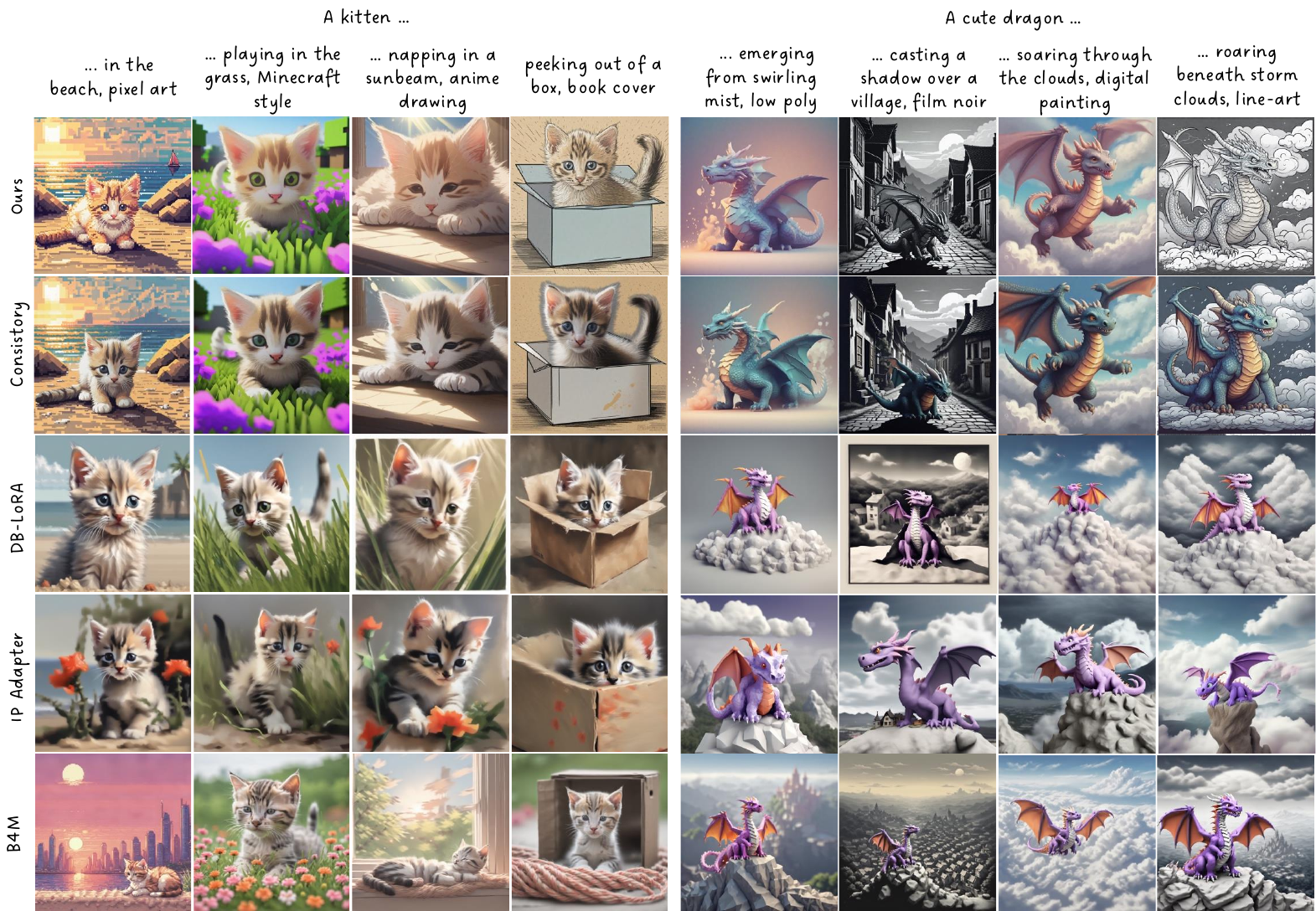}
    \caption{{Qualitative comparison of our method with Consistory, DB-LoRA, IP-Adapter and B4M demonstrates its effectiveness across varying text descriptions, character consistency, and style alignment. Unlike other methods, our approach preserves character features and maintains consistent appearance while faithfully adhering to the specified style and textual descriptions.}}
    \Description{A grid of images comparing results of our method to baselines.}
    
    \label{fig:comparison}
\end{figure*}

\begin{table}[t]
\centering
\caption{Symbol Definitions}
\label{tab:symbols}
\resizebox{\linewidth}{!}{
\begin{tabular}{@{}cl@{}}
\toprule
\textbf{Symbol} & \textbf{Description} \\
\midrule
$V_{\text{sd}, i}$ & self-attention Values of $i$-th image in the SDXL model \\
$B$ & the batch size \\
$d, H, W$ & dimension, height and width in the latent feature space \\
$N = HW$ & the number of total patches \\
$i$ & the index of $i$th image in the batch \\
$a$ & the anchor image index or indices \\
$Q_\alpha, K_\alpha, V_\alpha$ & self-attention queries, keys, values for sample $\alpha\in [B]$ \\
$h_\alpha$ & self-attention hidden state of image $\alpha$ \\
$z_\alpha$ & self-attention latent of image $\alpha$ \\
$s_\alpha$ & the subject indices in image $\alpha$ \\
$C_{\alpha'}(s_\alpha)$ & mapping of patches between image $\alpha$ and image $\alpha'$ \\
$\mathcal{A}$ & Adaptive Instance Normalization (AdaIn)\\
\bottomrule
\end{tabular}
}
\end{table}

\begin{table*}[h]
\centering
\caption{Comparing our method to other zero-shot methods that achieve style or identity consistency, each with its own distinct goal. The component modification part shows the modification of the presentation of each generated image, while the component import part shows how the self-attention of the model is modified to have a cross attention component. The methods in the table are Consistory \cite{tewel2024consistory}, Cross-Image Attention \cite{alaluf2023crossimage}, StyleAligned \cite{Hertz2023StyleAI} and IlluSign \cite{bruner2025illusign}. Our method is the only one that has an identity goal as well as a style goal, on top of the prompt faithfulness goal, which requires a much more nuanced solution. Prompt-to-Prompt \cite{hertz2022prompt} is not listed as its attention modifications are done in the cross-attention layer and not the self-attention layer.}
\label{tab:component-methods}
\resizebox{\textwidth}{!}{
\begin{tabular}{@{}lcccccc@{}}
\toprule
\textbf{Method} &
\multicolumn{4}{c}{\textbf{Component Modification}} &
\multicolumn{2}{c}{\textbf{Component Crossing}} \\
\cmidrule(l){2-5} \cmidrule(lr){6-7}
&  \textbf{h/z} & \textbf{Q} & \textbf{K} & \textbf{V} &\textbf{K} & \textbf{V} \\
\midrule
{Consistyle  (ours)} &
-- & {$Q_i[s_i] \leftarrow$  $Q_a[C_a(s_i)]$} &
{$K_i[s_i] \leftarrow$  $K_a[C_a(s_i)]$} &
{$V_i \leftarrow V_{\text{sd}, i}$} &
$K_i \leftarrow [K_i$, $K_j[s_j]]$ &
$V_i \leftarrow [V_i, \mathcal{A}(V_j[s_j], V_i)]$ \\

Consistory & 
$h_i[s_i] \leftarrow h_a[C_a(s_i)]$ & 
-- & 
-- & 
-- &
$K_i \leftarrow [K_i$, $K_j[s_j]]$ &
$V_i \leftarrow [V_i$, $V_j[s_j]]$ \\

{Cross-Image Atten.} & 
$z_i \leftarrow \mathcal{A}(z_i, z_a)$ & 
-- & 
$K_i \leftarrow K_a$ &
$V_i \leftarrow V_a$ &
-- &
--  \\

StyleAligned & 
-- & 
$Q_i \leftarrow \mathcal{A}(Q_i, Q_a)$ & 
$K_i \leftarrow \mathcal{A}(K_i, K_a)$ & 
-- &
$K_i \leftarrow [K_i, K_a]$ &
$V_i \leftarrow [V_i, V_a]$ \\

IlluSign & 
-- & 
$Q_i \leftarrow Q_i + \frac{1}{2} Q_a$ & 
$K_i \leftarrow K_a$ & 
$V_i \leftarrow V_a$ & 
-- &
--\\

\bottomrule
\end{tabular}
}
\end{table*}

\subsection{Transferring Style While Maintaining Appearance}\label{sec:attn_transfer}
To enhance the consistency of subjects across prompts, we focus on the selective transfer of keys and queries between subjects presented in the array of images at early stages, avoiding value exchange. This approach reduces the risk of unintended style propagation by preserving fine-grained details while maintaining subject structure and identity. To align with the subjects' structural differences, we use a correspondence mapping.

We aim to transfer style from the target image while adopting the semantic content of the source images. To achieve this, we first perform a vanilla pass through the SDXL model, during which we store the value matrices $V_{sd}$ from the self-attention layer of the decoder at the highest-resolution transformer block $[64\times64]$, and only during the early diffusion steps, specifically at steps $\frac{n}{10}$ to $3\frac{n}{10}$ (where $n$ is the number of steps). We also obtain subject masks $s_1, s_2, \dots$ using a threshold over the attention maps matching the subject token query in the cross-attention layer.

In the subsequent Consistyle pass, we apply the DIFT-based feature mapping between queries and keys, and inject the stored values at the corresponding layers and diffusion steps to:

\begin{equation}
    \begin{gathered}
        K_i[s_i] \leftarrow  K_a[C_a(s_i)] \\
        Q_i[s_i] \leftarrow  Q_a[C_a(s_i)] \\
        V = V_{sd}\,
    \end{gathered}
\end{equation}
\noindent where  $Q_i, K_i \in \mathbb{R}^{N \times h\times C}$, h are the attention heads and C number of features in the corresponding decoder layer $\in [32,64]$, a is the anchor image index(ices), and $C_a$ is the patch mapping induced from DIFT features \cite{Tang2023EmergentCF} of the patches computed during a previous run of the model between an anchor image and the target image, note that if there are multiple anchor images, the most similar patch across the anchors is used.

\subsection{AdaIN for Style Preservation in Attention Crossing}
\label{ref:adain}

As direct attention components injection can align the subject's details across images, in order to converge the subjects to the same structure we use an attention crossing component in which Queries may attend to Keys and Values of the images in the batch. Although it leads to improved consistency, the incorporation of Values between different images can lead to style contamination, as Values inherently carry fine-grained texture and color details.

Our approach mitigates this by applying adaptive instance normalization (AdaIN) \cite{adain} to the values before cross-subject attention, effectively isolating semantic content from style-specific features, since the statistical distribution of features is a key aspect of style, and texture is often defined in terms of such statistics \cite{Haralick}. By matching feature statistics using AdaIN \cite{adain}, we preserve the intended style: merely normalizing to constant values would distort the distribution and alter the resulting image style.

The $\mathcal{A}$ operator  operation is defined as follows,
\begin{equation}
    \mathcal{A}\bigl(x,\,y\bigr)
    = \sigma\!\left(y\right)
      \!\left(
        \frac{x-\mu\!\left(x\right)}{\sigma\!(x)}
      \right)
      +\mu\!\left(y\right)
\end{equation}
where \(\mu(\cdot)\) and \(\sigma(\cdot)\) denote the mean and standard deviation functions, respectively. The subsequent attention crossing is thus, for $i\!\in\![B]$
\begin{gather}
\label{equation:attention_crossing}
  V_i \leftarrow \left[\operatorname{AdaIN}\bigl(V_1[s_1], V_i\bigr), \cdots, V_i, \cdots, \operatorname{AdaIN}\bigl(V_n[s_n], V_i\bigr) \right], \\
    K_i \leftarrow \left[K_1[s_1], \cdots, K_i, \cdots, K_n[s_n]\right]\,,
\end{gather}
\noindent where $K_l, V_l$ are the keys and values in the self-attention layer matching image $l$, and $s_l$ are the mask indices of the subjects in the image.

\subsection{Summarizing the differences from other methods}

With a clear view of the method and the associated terminology, we revisit the comparison to previous work, now on a clear technical level. A comparison to the most similar contributions can be found in Table ~\ref{tab:component-methods}. The table separates the modification of the self-attention mechanism from the step of selectively importing content from other attention maps, creating a cross-attention scheme. 
We observe notable differences in the handling of self-attention component imports and modifications across various methods. Most approaches rely on direct Value imports from the anchor image. For instance, StyleAligned and Consistory both utilize direct Value imports, with Consistory further incorporating hidden state injections for enhanced visual consistency. Cross-Image Attention \cite{alaluf2023crossimage} similarly relies on direct Value modification, emphasizing precise texture and color transfer.

In contrast, our approach avoids direct Value imports to prevent the style misalignment typically associated with direct appearance transfer. Instead, we employ targeted Value modifications, aligning values to the original $SDXL$ features $V_{sd}$ for improved style alignment. Additionally, we apply AdaIN to regulate the statistical properties of imported Values, ensuring smoother integration into the target domain. Unlike Consistory and Cross-Image Attention, we do not use the hidden values of the feature embedding for the self-attention intervention, and instead modify $Q$, $K$, and $V$ selectively, giving $V$ a different treatment. StyleAligned modified only $Q$ and $K$, using AdaIn that in our method is applied for the $V$ of the dictionary expansion part (the crossing of attentions).

There are also several differences, which are not captured in Table ~\ref{tab:component-methods}. For example, the timing of the attention intervention varies across methods. Key and Query modifications as well as the $V_{sd}$ injection in Consistyle, are confined to early generation steps, while other modifications and imports are applied across all time steps as specified in Sec. ~\ref{sec:attn_transfer}. Also, the Keys and Queries modification is done only on the text-guided part of the batch in the self-attention layer, where the $V_{sd}$ injection and attention crossing components are also applied on the non-guided elements, which seems to yield a slight improvement.

\begin{figure}[t] %
    \centering
    \includegraphics[width=0.5\textwidth]{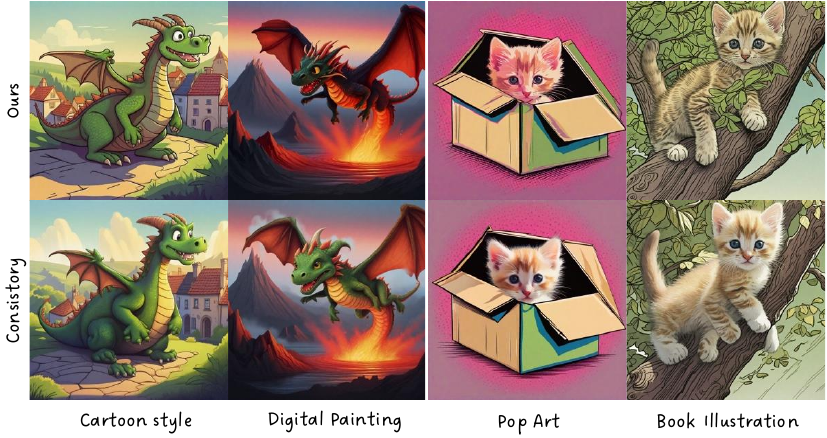}
    \caption{\textit{Harmonization.} Our method preserves the desired style, seamlessly integrating characters into stylized contexts such as cartoons or illustrations. It adapts both the appearance and the setting, e.g., casting firelit shadows on a dragon or applying a pinkish tone to a kitten in a similar environment.}
    \Description{Image harmonization.}
    
    \label{fig:harmonizarion}
\end{figure}

\section{Experiments}

We evaluate our method using three core metrics: prompt alignment, consistency, and style alignment, comparing it against four state-of-the-art (SOTA) baselines designed for consistent image generation. The first baseline is a training-free Consistory approach \cite{tewel2024consistory}, the second is an encoder-based method, IP-Adapter \cite{ye2023ip-adapter}, and the third and fourth are personalization training-based methods, DreamBooth-LoRA (DB-LoRA) \cite{ryu2023lora} and Break-for-Make (B4M) \cite{xu2025b4m}. For training details of our baseline models, please refer to the supplementary material. Unlike Consistory, which directly modifies the diffusion process, both IP-Adapter and DB-LoRA utilize a single reference image for personalization.

\subsection{Implementation Details and Latency}
Our method was evaluated on an A100 GPU (40GB) with a batch size of 5. SDXL and IPAdapter requires ~12 sec per image. Consistory involves two passes, amounting to 24 sec per image, while our method requires three passes, totaling 36 sec per image.
In contrast, training-based baselines involve substantially higher computational overhead. DB-LoRA requires around 10 minutes of training per subject. B4M entails 70 min per subject, 70 min per style, and an additional 70 min for each subject–style combination, resulting in 210 min for a single subject–style pair.

\subsection{Qualitative Results}

\begin{figure*}[hb]
    \centering
    \includegraphics[width=0.78\textwidth]{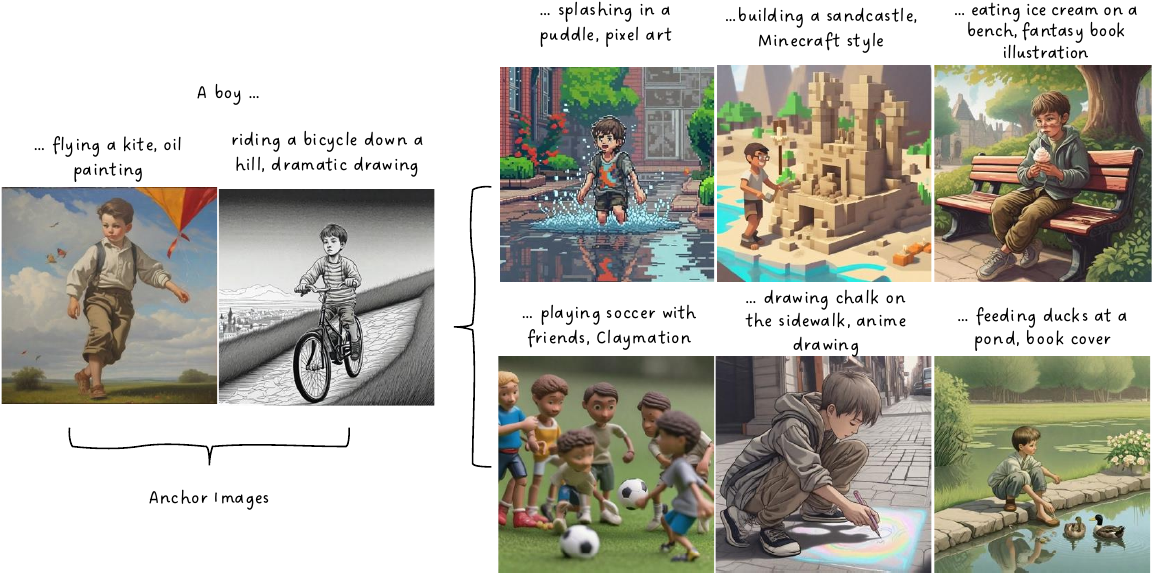}
    \vspace{1em}
    
    \centering
    \includegraphics[width=0.78\textwidth]{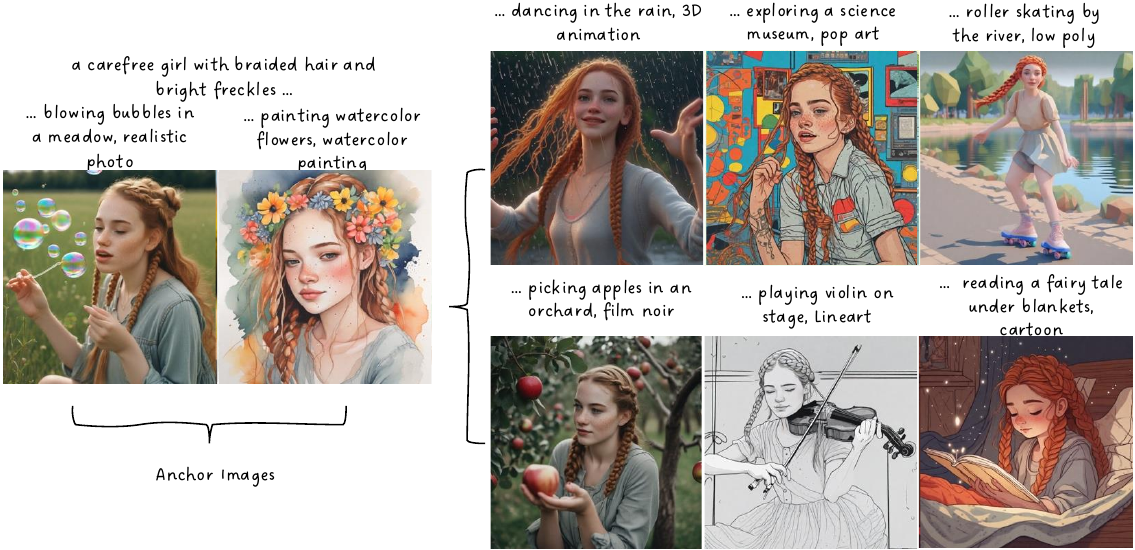}
    \vspace{0.5em}
    \centering
    \includegraphics[width=0.78\textwidth]{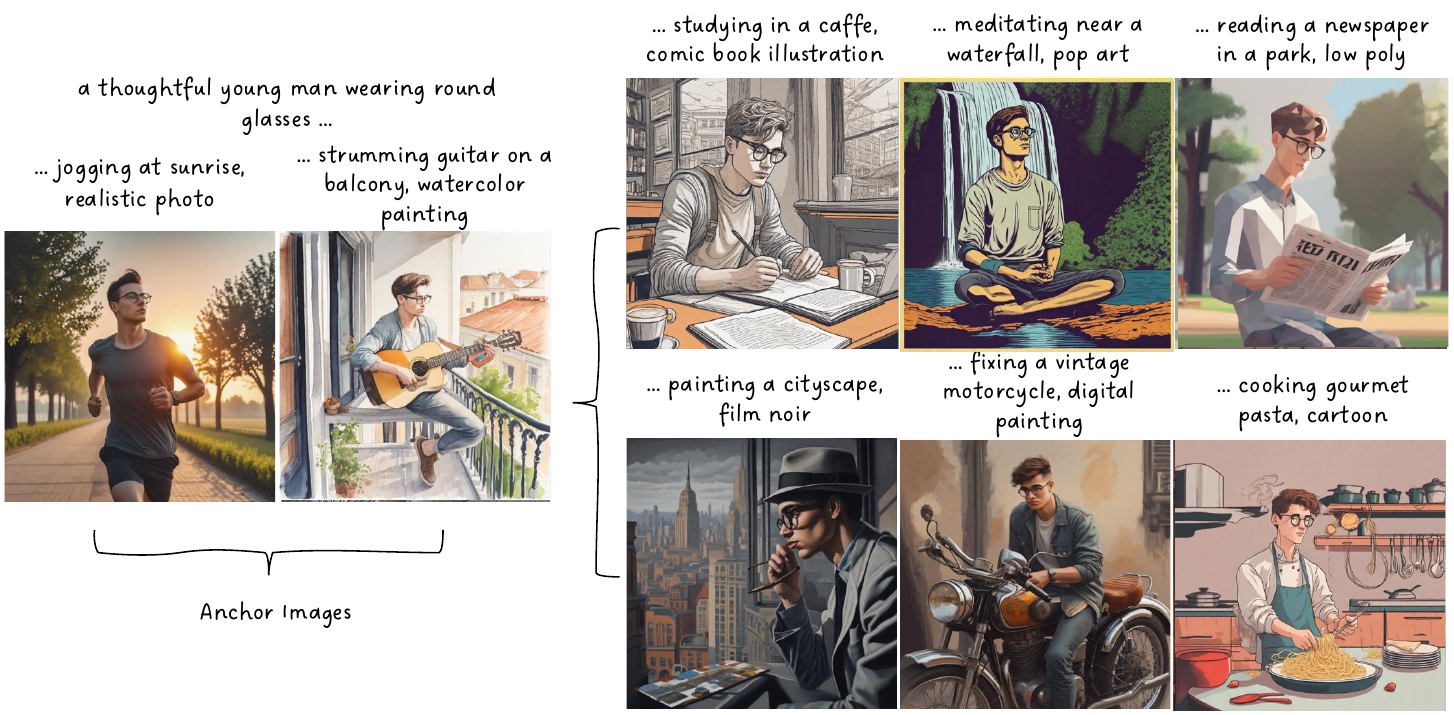}
    \caption{Qualitative results highlighting the consistency, style alignment, and textual coherence of our method, guided by two anchor images.}

    \label{fig:examples_page_1}
\end{figure*}

Our method is designed to accommodate a wide range of visual styles, including highly detailed photographic renderings, abstract illustrations and 3D aesthetics. In Fig. ~\ref{fig:comparison} and Fig. ~\ref{fig:qulitative_part_2} we illustrate its ability to preserve style, maintain character consistency, and align with textual descriptions. 
As shown, B4M, IP-Adapter, and DB-LoRA exhibit strong character consistency, yet they fail to respect the intended style and often feature very little variation. Notably, in both subjects these three methods render the kitten using a realistic style and the dragon with a 3D animation style across all images, disregarding the specified stylistic variations due to overfitting to the reference image.
In Fig.~\ref{fig:qulitative_part_2}, B4M shows better alignment to the style but the other two baselines fail in this. Moreover, we observe that LoRA-based methods tend to suffer from overfitting and require an exhaustive hyperparameter search grid to maintain both content and style.
Consistory demonstrates strong character consistency and faithful prompt alignment; however, it frequently fails to harmonize the subject with the stylistic setting. In Fig. ~\ref{fig:comparison}, this issue is evident in the dragon images for the film noir and line-art styles—both intended to be black and white, yet Consistory generates colored images or subjects. This misalignment extends beyond color. For example, in the kitten images (first and fourth row), the subject exhibits a highly realistic texture that clashes with the surrounding stylistic context. Furthermore, as shown in Fig. ~\ref{fig:harmonizarion}, although the compositions may initially seem coherent, closer examination reveals integration flaws—characters often appear visually `stitched' onto the background rather than naturally embedded within the scene. In contrast, our approach maintains both consistency and stylistic harmony, allowing characters to seamlessly integrate into their environments and preserving the intended aesthetic.
The application of our method to multiple anchor images is demonstrated in Fig.~\ref{fig:examples_page_1}.

\subsection{Quantitative Results}
\begin{table*}[t]
\centering
\caption{Comparison of various methods along perceptual, content, and style metrics. (Mean ± SD)}
\label{tab:numerical-results}
\begin{tabular}{lccccccc}
\toprule
\textbf{Method} & \textbf{DreamSim $\downarrow$} & \textbf{CLIPScore $\uparrow$} & \textbf{CLIPScore, Styled $\uparrow$} & \textbf{LPIPS $\downarrow$} & \textbf{Gram L2 $\downarrow$} & \textbf{DINO $\uparrow$}\\
\midrule
\multicolumn{7}{l}{\textbf{Full Dataset (500 samples)}} \\ 
Consistyle (ours)   & $0.40 \pm 0.10$ & $\mathbf{32.84 \pm 1.66}$ & $\mathbf{36.03 \pm 1.69}$ & $\mathbf{0.21 \pm 0.06}$ & $\mathbf{1.25 \pm 0.69}$  & $\mathbf{0.85 \pm 0.08}$\\
Consistory   & $0.33 \pm 0.12$ & $32.75 \pm 1.53$ & $35.34 \pm 1.67$ & $0.27 \pm 0.07$ & $1.85 \pm 1.13$  & $0.78 \pm 0.11$\\
IP Adapter   & $\mathbf{0.25 \pm 0.08}$ & $30.98 \pm 2.03$ & $32.07 \pm 2.08$ & $0.43 \pm 0.07$ & $2.96 \pm 1.57$  & $0.54 \pm 0.17$\\
DreamBooth-LoRA      & $0.28 \pm 0.13$ & $32.43 \pm 1.74$ & $34.33 \pm 2.20$ & $0.42 \pm 0.07$ & $2.60 \pm 1.32$  & $0.59 \pm 0.15$\\
Cross-Img            & $0.55 \pm 0.12$ & $30.68 \pm 1.36$ & $33.32 \pm 1.52$ & $0.33 \pm 0.02$ & $1.43 \pm 0.70$  & $0.85 \pm 0.08$\\
IlluSign & $0.53 \pm 0.12$ & $30.98 \pm 1.48$ & $33.82 \pm 1.31$ & $0.38 \pm 0.02$ & $1.40 \pm 0.67$  & $0.82 \pm 0.08$\\
\midrule
\multicolumn{7}{l}{\textbf{Subset (100 samples)}} \\
Consistyle (ours)   & $0.46 \pm 0.12$ & $\mathbf{33.00 \pm 1.60}$ & $\mathbf{36.55 \pm 1.18}$ & $\mathbf{0.30 \pm 0.04}$  & $\mathbf{1.35 \pm 0.9}$ & $\mathbf{0.86 \pm 0.08}$ \\
B4M   & $\mathbf{0.38 \pm 0.17}$ & $30.67 \pm 2.03$ & $33.30 \pm 1.60$ & $0.72 \pm 0.02$ \  & $2.75 \pm 1.71$  & $0.55 \pm 0.15$ \\
\bottomrule
\end{tabular}
\end{table*}

To evaluate our approach, we constructed a dataset of 25 prompt groups, each consisting of a subject description paired with ten distinct prompts. The subjects were divided into four categories: humans, animals, fantasy creatures, and inanimate objects, generated using ChatGPT. In addition, we curated two style groups, each containing ten diverse styles obtained from an online resource \cite{agsteiner2023styles}.
By combining each prompt group with each style group, we obtained 500 images, organized into 50 unique sets. Each set corresponds to one experiment and consists of a batch of $B=10$ images, where both prompts and styles vary, ensuring that no prompt or style is repeated within the same batch.

For our evaluation we employ several automated metrics following prior work \cite{tewel2024consistory, gao2024styleshot, avrahami2024chosen}. Text alignment is measured using CLIPScore \cite{hessel-etal-2021-clipscore}, both with and without style descriptions. Subject consistency is evaluated using DreamSim \cite{fu2023dreamsim} as proposed by \cite{tewel2024consistory} with background removal. 
For style alignment, we measure Gram Matrix distance \cite{gatys2015neural}, and for perceptual similarity we measure LPIPS \cite{zhang2018perceptual}, and DINO similarity \cite{oquab2023dinov2}.
These latter metrics (LPIPS and DINO) are used to evaluate content fidelity between stylized images \cite{xu2025b4m,wang2025sigstyle}, by comparing each stylized output to vanilla SDXL generations that serve as style references.

We conduct two sets of experiments: one using our full dataset of 500 images, and another using a subset of 100 images. 
The latter was necessary due to the computational cost of training B4M, which requires approximately four hours per image.
The results of these experiments are presented in Table~\ref{tab:numerical-results}. As can be seen, our method outperforms all baselines in the prompt alignment and style alignment scores. However, we note a critical limitation in automated metrics for consistency when operating in style-diverse settings. Automated methods tend to over-penalize stylistic variation, potentially rewarding overly consistent outputs that lack stylistic diversity.

\subsection{User Study}
\begin{table}[t]
\centering
\caption{{User study results showing pairwise preference percentages across three criteria. Each pair was rated for style alignment, subject consistency, and text alignment. Tie votes are split equally.}}
\label{tab:user_study_summary}
\begin{tabular}{lcccc}
\toprule
\textbf{Question} & \textbf{Method A} & \textbf{Method B} & \textbf{A \%} & \textbf{B \%} \\
\midrule
Style        & DB LoRA     & Ours         & 25.2\% & \textbf{74.8\%} \\
Subject  & DB LoRA     & Ours         & \textbf{64.5\%} & 35.5\% \\
Text         & DB LoRA     & Ours         & 39.7\% & \textbf{60.3\%} \\
\midrule
Style        & Consistory  & Ours         & 15.1 \% & \textbf{84.8\%} \\
Subject  & Consistory  & Ours         & 46.8\% & \textbf{53.2\%} \\
Text         & Consistory  & Ours         & 40.0\% & \textbf{60.0\%} \\
\midrule
Style        & DB LoRA     & Consistory   & 21.3\% & \textbf{78.7\%} \\
Subject  & DB LoRA     & Consistory   & \textbf{67.2\%} & 32.8\% \\
Text         & DB LoRA     & Consistory   & 29.4\% & \textbf{70.6\%} \\
\bottomrule
\end{tabular}
\end{table}
Since automatic metrics only partly correlated with human perception, especially when measuring subject consistency when varying style, we conducted a user study.
The user study evaluates human preferences regarding style alignment, text alignment, and subject consistency. Each user is exposed to 12 random prompt and style combination. In each, the users are presented with pair of images generated by three methods: Consistory, Consistyle, and DB-LoRA. For each pairwise comparison participants answered three questions, one for each criterion. Users had the option to vote in favor of one set of images or to indicate that both methods performed equally well; in such cases, each method received half a vote. See supplementary material for more details.

The results are depicted in Table ~\ref{tab:user_study_summary}. As can be seen, our method outperforms the baselines in both style alignment and text alignment, and demonstrates higher subject consistency than the Consistory method. Meanwhile, DB-LoRA was preferred for subject consistency, as it maintains strong consistency across generations, although it often ignores style and textual descriptions, resulting in the lowest scores for those criteria.

\subsection{Ablations}
To better understand the impact of key design choices in our technique, we conduct a series of ablation studies, each isolating a critical component to assess its influence on consistency, style alignment, and overall image quality. The studied variants are:

\begin{enumerate}[label=(\roman*)]
    \item Consistory with an increased step budget (75 steps) to equate the runtime to our method,
    \item Consistyle without query injection,
    \item Consistyle without key injection,
    \item Consistyle without both query and key injection,
    \item Consistyle with direct identity injection between images instead of subject-based DIFT mapping,
    \item Consistyle without AdaIN in the attention crossing.
\end{enumerate}

\begin{figure*}[t]
    \centering
    \includegraphics[width=0.8\textwidth]{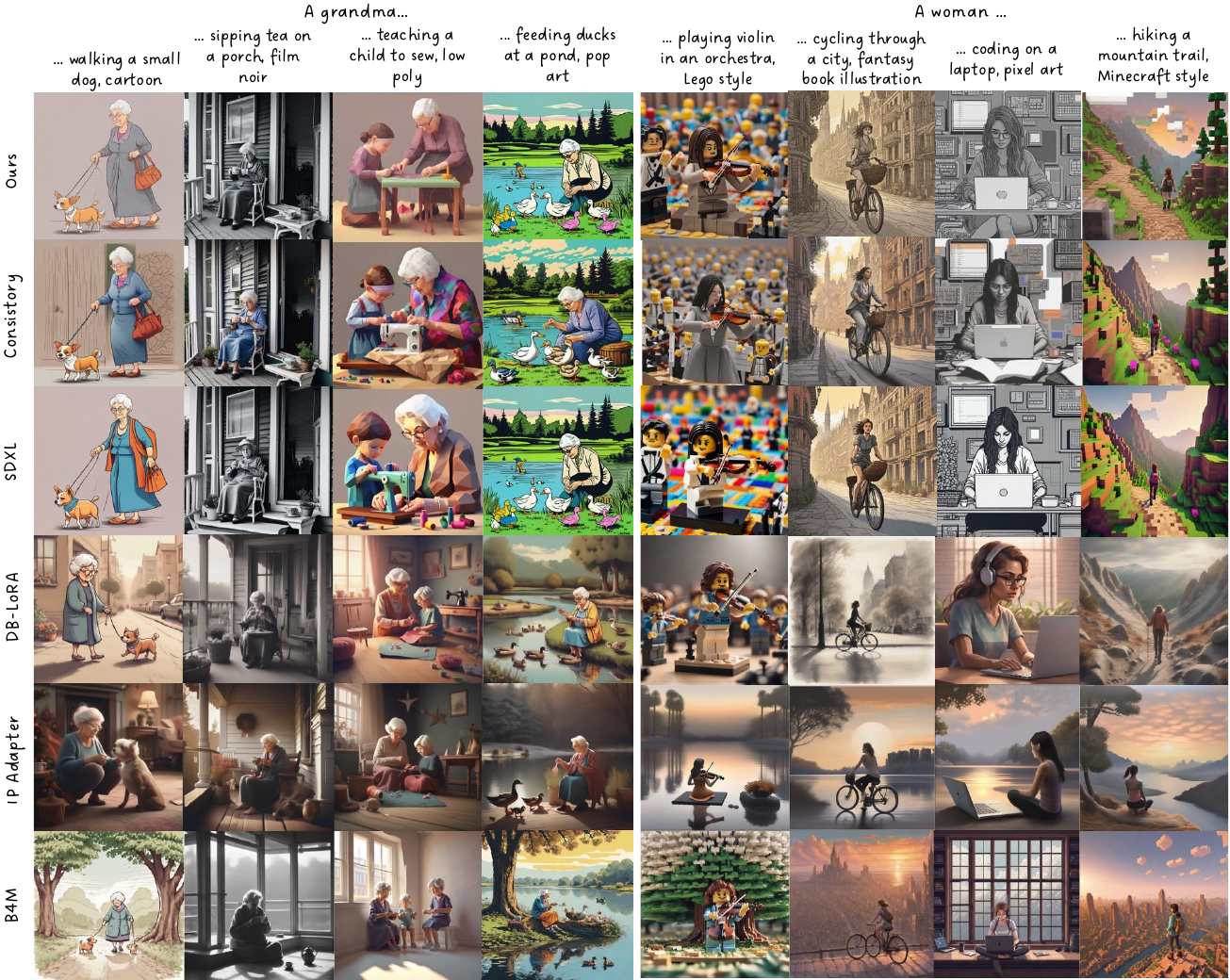}
    \vspace{-1em}
    \caption{Additional comparison.}
    \label{fig:qulitative_part_2}


        
        
        \noindent
    \raisebox{-27.3em}[0pt][0pt]{%
    \begin{minipage}[t]{0.33\textwidth}
        \includegraphics[width=\textwidth]{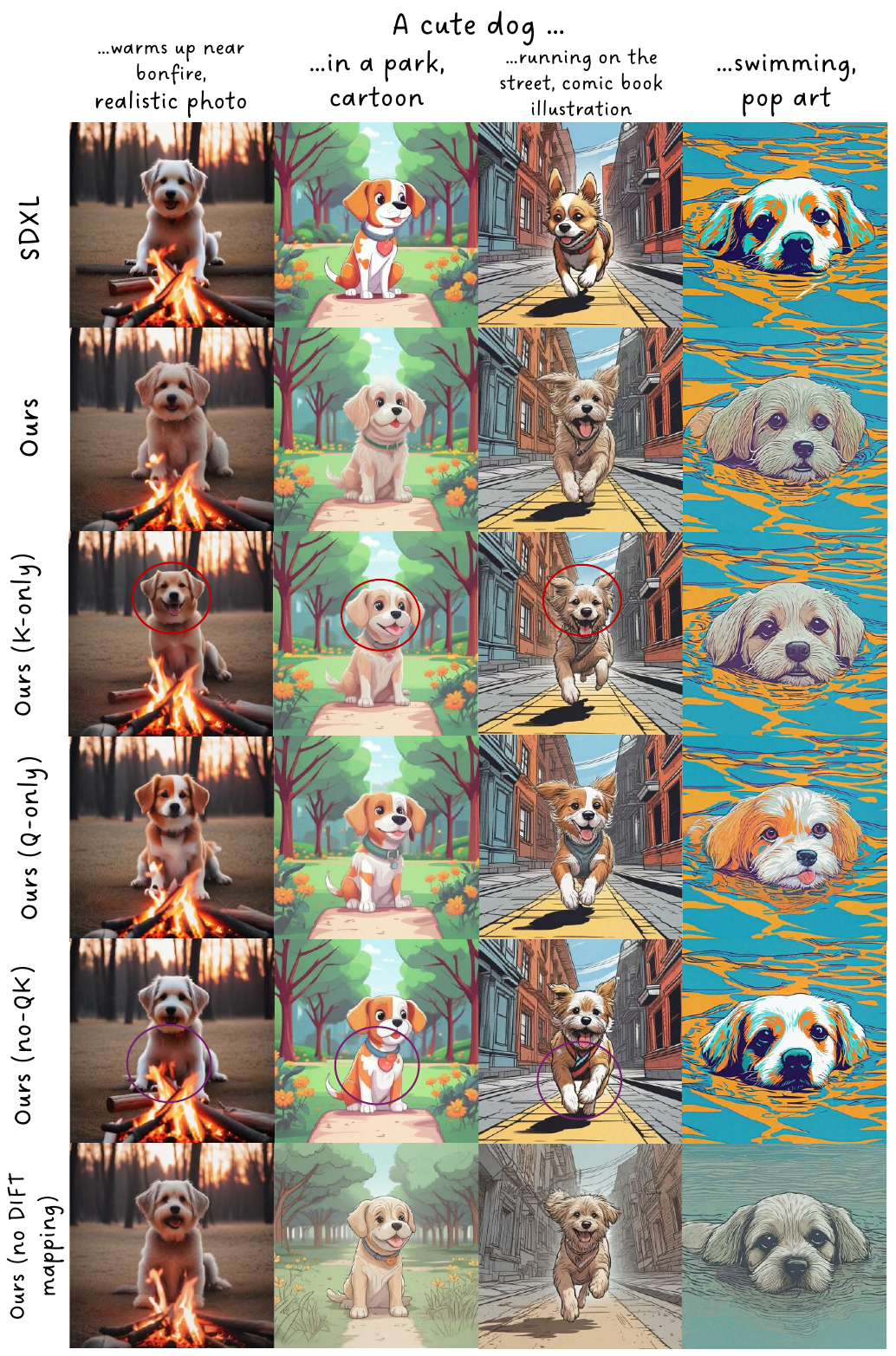}
        
        \vspace{-1.2em}
        
        \parbox{1.2\textwidth}{%
            \captionof{figure}{Ablation study of the attention transfer component. We compare our full method vs variations with partial or no transfer of Keys and Queries. Inconsistent details such as color mismatch are marked.}
            \label{fig:ablations_attn_transfer}
        }
    \end{minipage}%
    }
    \hspace{4em}
    \raisebox{-25em}[0pt][0pt]{ 
        \begin{minipage}[t]{0.35\textwidth}
            \parbox[t]{\textwidth}{
                \includegraphics[width=\textwidth]{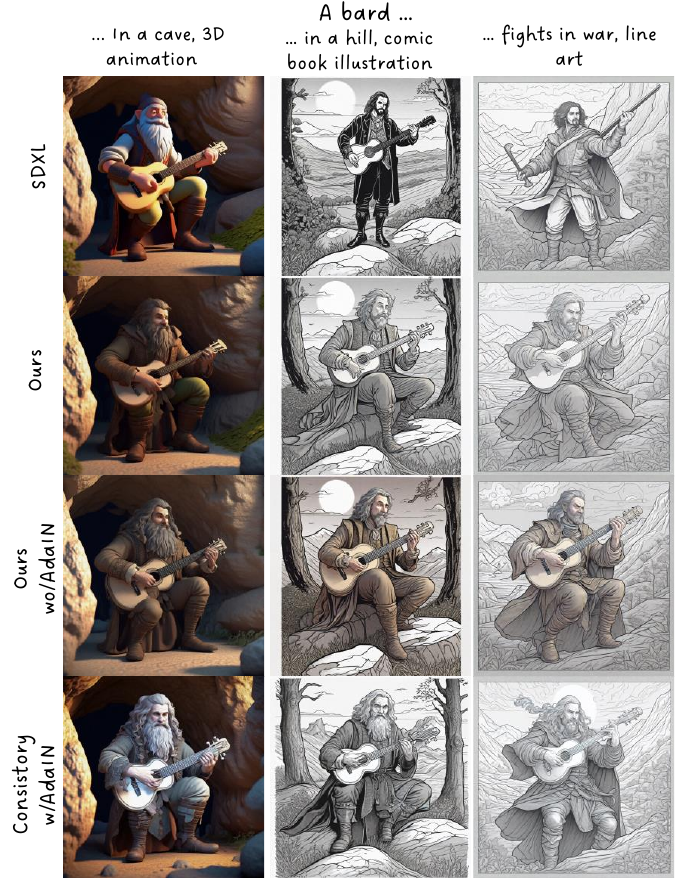}
                \captionof{figure}{Effect of AdaIN on style alignment. We compare results across original outputs, Consistyle, Consistyle without AdaIN, and Consistory. Notice the color shifts, especially when AdaIN is removed or Consistory is used.}
                \label{fig:ablations_adain}
            }
        \end{minipage}
    }
\end{figure*}

The quantitative outcomes are summarized in Tab.~\ref{tab:numerical-results-ablations}, with representative examples shown in Fig.~\ref{fig:ablations_attn_transfer} and Fig.~\ref{fig:ablations_adain}. Variant (i) demonstrates that extending Consistory’s runtime does not substantially change its performance: our method continues to surpass it across all metrics except DreamSim, consistent with earlier results. Variants (ii)--(v) confirm the importance of the attention injection design. Removing the query, key, or both worsens similarity scores, since the attention injection mechanism is designed to enhance subject consistency. At the same time, its absence slightly improves some style-alignment metrics, as fewer details are transferred across images. 

In variant (v), we replace the DIFT-based mapping with a full direct identity injection from an anchor image.
Since no patch-based mapping is available and subject segmentations vary in size, we inject keys and values from the entire image. In addition, since no distance measures are available to compare across different anchors, the injection is restricted to the first image in the batch.
As shown in Tab.~\ref{tab:numerical-results-ablations} and illustrated in 
Fig.~\ref{fig:ablations_attn_transfer}, this variant leads to feature leakage  across non-subject regions, resulting in degraded style-alignment scores. In  particular, the image backgrounds shift noticeably in their tone compared to their original SDXL counterparts, unlike other variants where the backgrounds remain nearly unchanged.

Finally, variant (vi) highlights the role of AdaIN: without it, values are transferred directly across images, which weakens style alignment, as reflected by a pronounced drop in the Gram metric and visible artifacts in Fig.~\ref{fig:ablations_adain}.  

Overall, all ablated versions perform worse than our full model, underscoring the necessity of each component in achieving both strong consistency and style fidelity.

\begin{table*}[t]
\centering
\caption{Comparison results of the ablation study. Bold values denote the best results and underlined values indicate the second best.(Mean ± SD)}
\label{tab:numerical-results-ablations}
\begin{tabular}{lcccccc}
\toprule
\textbf{Ablation Method} & \textbf{DreamSim $\downarrow$} & \textbf{CLIPScore $\uparrow$} & \textbf{CLIPScore, Styled $\uparrow$} & \textbf{LPIPS $\downarrow$} & \textbf{Gram L2 $\downarrow$} & \textbf{DINO $\uparrow$} \\

\midrule
Consistory (original, 50 steps)   & $\mathbf{0.33 \pm 0.12}$ & $32.75 \pm 1.53$ & $35.34 \pm 1.67$ & $0.27 \pm 0.07$ & $1.85 \pm 1.13$  & $0.78 \pm 0.11$\\
(i) Consistory (75 steps) & $0.40 \pm 0.12$ & $32.79 \pm 1.55$ & $35.40 \pm 1.67$ & $0.42 \pm 0.04$ & $1.57 \pm 0.30$ & $0.77 \pm 0.11$ \\
\midrule
Consistyle (full)   & $\underline{0.40 \pm 0.10}$ & $\underline{32.84 \pm 1.66}$ & $\mathbf{36.03 \pm 1.69}$ & \underline{$0.21 \pm 0.06$} & $1.25 \pm 0.69$  & $0.85 \pm 0.08$ \\
(ii) Consistyle (no Q injection) & $0.48 \pm 0.11$ & $32.75 \pm 1.64$ & $\underline{35.99 \pm 1.61}$ & $0.28 \pm 0.03$ & $\underline{1.18 \pm 0.23}$ & $\underline{0.87 \pm 0.08}$ \\
(iii) Consistyle (no K injection) & $0.45 \pm 0.12$ & $32.73 \pm 1.59$ & $35.94 \pm 1.71$ & $0.32 \pm 0.04$ & $1.23 \pm 0.25$ & $0.84 \pm 0.08$ \\
(iv) Consistyle (no QK injection) & $0.48 \pm 0.11$ & $32.60 \pm 1.66$ & $35.80 \pm 1.65$ & $\mathbf{0.18 \pm 0.03}$ & $\mathbf{0.72 \pm 0.11}$ & $\mathbf{0.94 \pm 0.05}$ \\
(v) Consistyle (w/o DIFT) & $0.43 \pm 0.12$ & $\mathbf{32.88 \pm 1.52}$ & $35.74 \pm 1.57$ & $0.49 \pm 0.03$ & $2.19 \pm 1.41$ & $0.74 \pm 0.14$ \\
(vi) Consistyle (no AdaIN) & $0.42 \pm 0.12$ & $32.81 \pm 1.59$ & $35.66 \pm 1.71$ & $0.23 \pm 0.03$ & $1.39 \pm 0.27$ & $0.82 \pm 0.09$ \\

\bottomrule
\end{tabular}
\end{table*}

\begin{figure}[h] %
    \centering
    \includegraphics[trim=10mm 0 0 0, clip,width=0.4\textwidth]{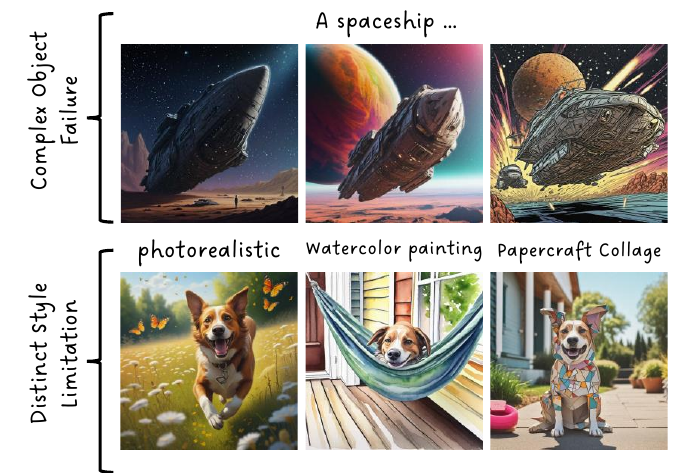}
    \caption{Demonstration of the method's limitations. The first row illustrates inconsistencies in generating a complex object (spaceship), where high visual detail leads to variation across images. The second row highlights a failure to align with a distinct style-specifically, the Papercraft Collage style, evident in the face details.}
    \Description{Two rows of generated images. The first row shows inconsistent depictions of a spaceship across images, reflecting difficulty with high-detail subjects. The second row attempts Papercraft Collage-style renderings but the face details seems realistic.}
    \label{fig:limitations}
\end{figure}

\section{Limitations}
Our approach has several limitations, as can be seen in Fig.~\ref{fig:limitations}. First, similar to Consistory \cite{tewel2024consistory}, it can produce suboptimal results when the correspondence mappings or cross-attention masks fail to accurately capture the intended relationships between images. 
Second, consistency can degrade for subjects with complex structures, such as large vehicles like boats and spaceships, where fine details are often challenging to maintain. This effect can extend to human subjects, where intricate facial features or clothing elements may vary across generations. The phenomenon is more prevalent when the initial subject interpretations differ significantly between images, such as generating a traditional wooden ship in one image and a modern yacht in another for the same "boat" prompt. While this issue can sometimes be mitigated by selecting different seeds, it remains a potential weakness in cases where precise subject alignment is critical. Finally, highly distinctive artistic styles, such as Papercraft Collage, Voxel Art or other niche 3D aesthetics, can occasionally lead to style misalignment, as these styles often introduce unique structural deformations or exaggerated textures that challenge the statistical alignment technique we employ. 

We also observed artifacts in some generated images. To verify that these do not originate from our method, we manually labeled 400 samples as having either minor artifacts (e.g., small visual glitches) or severe ones (e.g., extra limbs). We found that minor artifacts occurred in $18.8\%$ of SDXL images, $20.0\%$ of ours, and $23.2\%$ of Consistory; major artifacts appeared in $2.0\%$, $2.2\%$, and $4.0\%$ respectively, which indicates that this stems from the base model.

\section{Conclusion}
We present Consistyle, a training-free approach for improving consistency while preserving style alignment in text-to-image generation. Our method leverages attention manipulation to enable controlled characteristic sharing between images, even in cases with significant appearance differences. This demonstrates that consistent image generation is feasible in style-diverse contexts, despite the typical entanglement of style and content that often challenges.

\begin{acks}
This work was supported by a grant from the Tel Aviv University Center for AI and Data Science (TAD). 
We would like to thank Yoad Tewel for his insightful feedback.
\end{acks}
\clearpage
\balance
\bibliographystyle{ACM-Reference-Format}
\bibliography{bibliography}

\clearpage

\clearpage
\appendix
\setcounter{page}{1}
\textbf{APPENDIX}
\section{Training Details for DB LoRA and B4M}
In this section, we provide a detailed overview of the training procedures used for the baselines we trained—DreamBooth LoRA (DB LoRA) and Break for Make (B4M).
\subsection{DreamBooth LoRA}
\paragraph{Workflow.} Instance images were generated with the first prompt of the prompt group, combined with the appropriate style from the style group. 
It was paired with class images generated with random seeds on the concept token to improve results while mitigating overfitting.
Instance prompts were constructed by substituting the concept token with a unique placeholder token.
\paragraph{Parameter Details.} Models were trained for $250$ steps at a resolution of $1024 \times 1024$, using a batch size of $1$ with gradient accumulation over $3$ steps. Mixed precision (\texttt{fp16}) and gradient checkpointing were applied to reduce memory usage. Optimization employed 8-bit Adam with a constant learning rate of $1 \times 10^{-4}$ and no warmup. SNR weighting was applied with $\gamma = 5.0$, and seeds were clamped to the 32-bit integer range for reproducibility.

\subsection{Break for Make}
The training procedure for B4M consists of three phases:  

\paragraph{Phase 1: Content LoRA}  
We first train a LoRA for the content reference (e.g., "a kitten"). For stable results, it is recommended to use at least three images. To this end, we generate one image with vanilla SDXL and two additional images with IP-Adapter in a "photo-realistic" style. These prompts are deliberately different from those used in our evaluation templates.  
To disentangle content from style, we additionally provide three style reference images, as suggested by the original authors, chosen from style categories not included in our evaluation set. Training is performed for 1000 steps, as recommended. However, we observe that overfitting may occur; in some cases, reducing the training to 500 steps yields better results.  

\paragraph{Phase 2: Style LoRA}  
Next, we train a LoRA for style references. We generate three images (e.g., landscape, cat, etc.) with a specific style description (e.g., pixel art, line-art). As in Phase 1, we train for 1000 steps and provide three additional content images to encourage disentanglement between content and style.  

\paragraph{Phase 3: Content–Style LoRAs}  
In the final phase, we train separate LoRAs for each content–style pair, again for 1000 steps. 

\paragraph{Inference}
At inference time, images are generated by composing the learned tokens corresponding to content and style. For example, a prompt could be: *“snq woman coding on a laptop, w@z pixel art style.”*

\section{User Study Instructions and Questions}
\label{app:userstudy}
\begin{figure*}[hb]
    \includegraphics[width=0.8\textwidth]
    {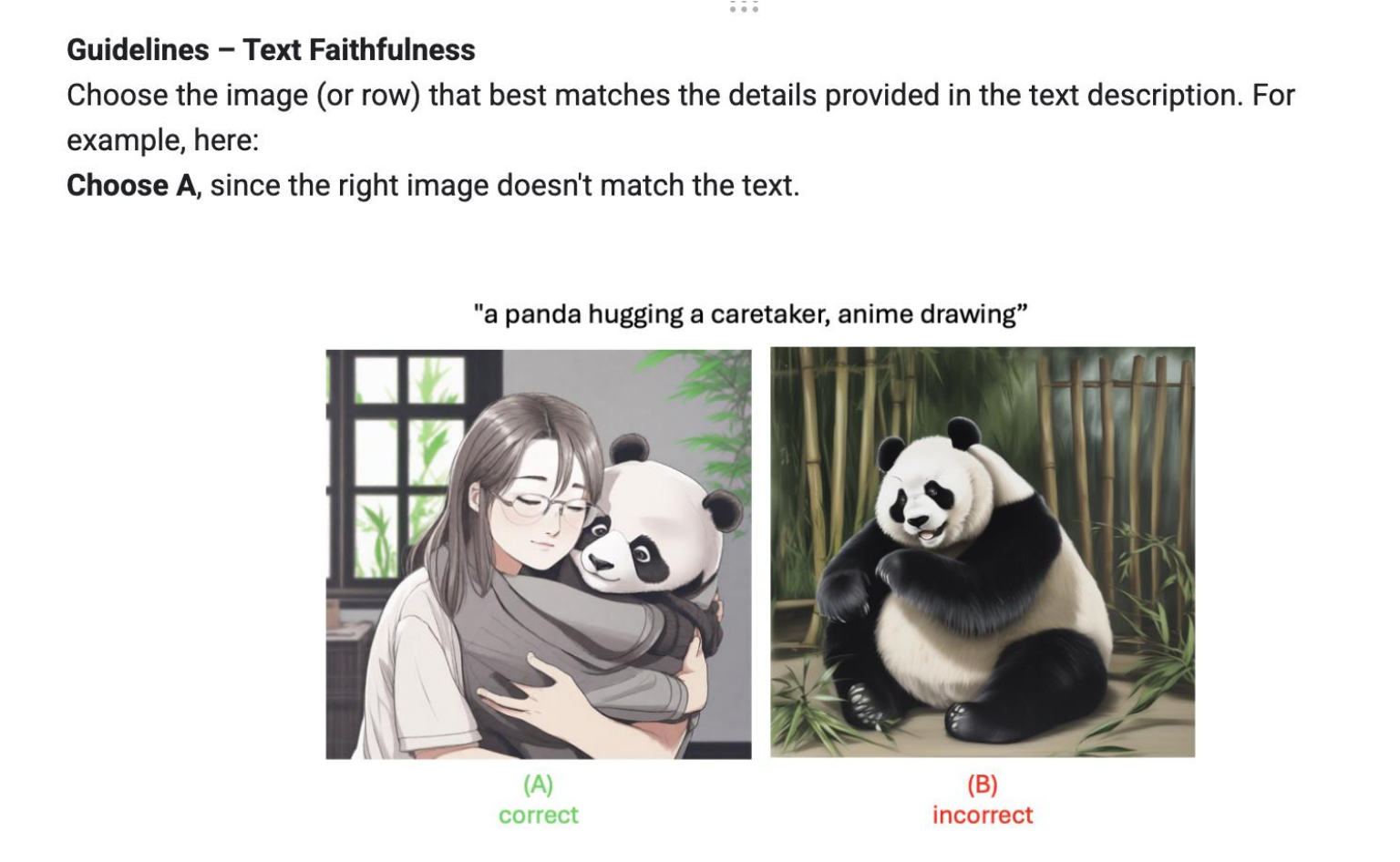}
    \caption{Guidelines from our user study.}
    \label{fig:user_study_struct2}
\end{figure*}

In the user study, participants were asked to choose one set of images per method. Before answering any questions, they were instructed to carefully read the guidelines, which explained the relevant terminology and outlined what to look for when evaluating subject identity, text alignment, and style alignment. For each task, an example was provided to illustrate the evaluation criteria. The full set of guidelines is shown in Fig.~\ref{fig:user_study_struct2}.

\begin{figure*}[hb]
    \centering
    \includegraphics[width=0.8\textwidth]{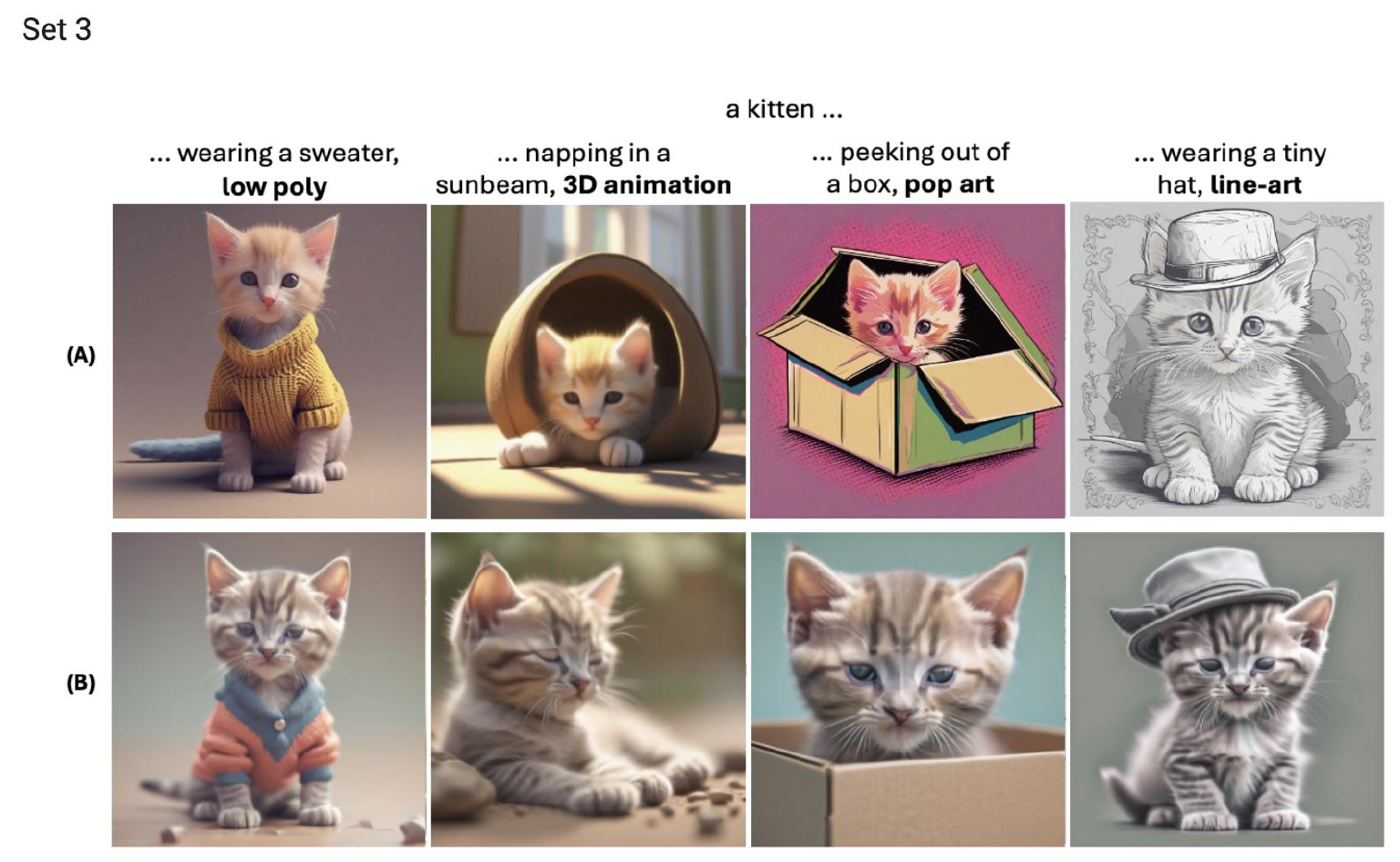}
    \caption{An example set of images used for comparison}

    \label{fig:user_study_example}
\end{figure*}

\begin{figure*}[hb]
    \centering
    \includegraphics[width=0.9\textwidth]{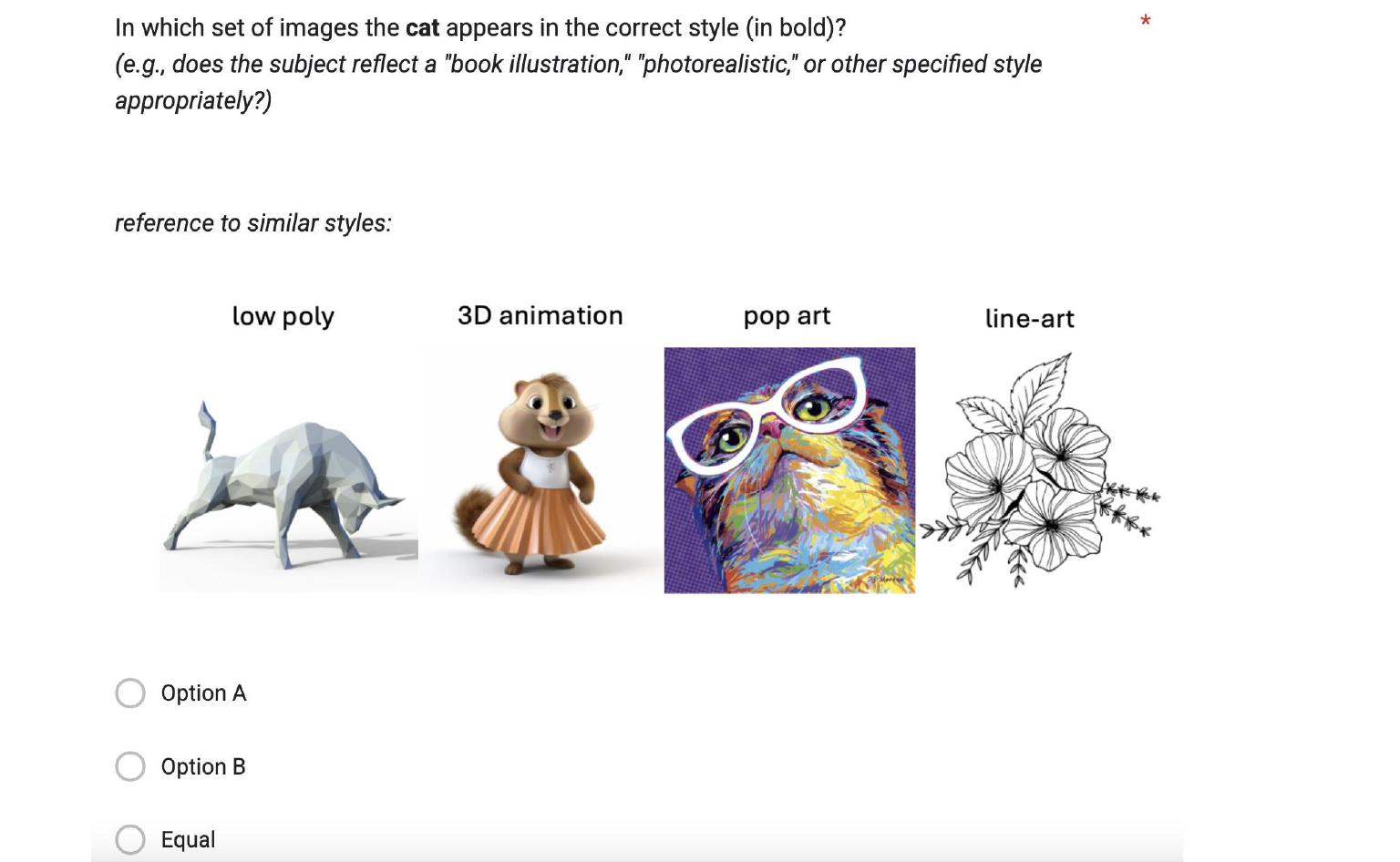}
    \centering
    \includegraphics[width=0.9\textwidth]{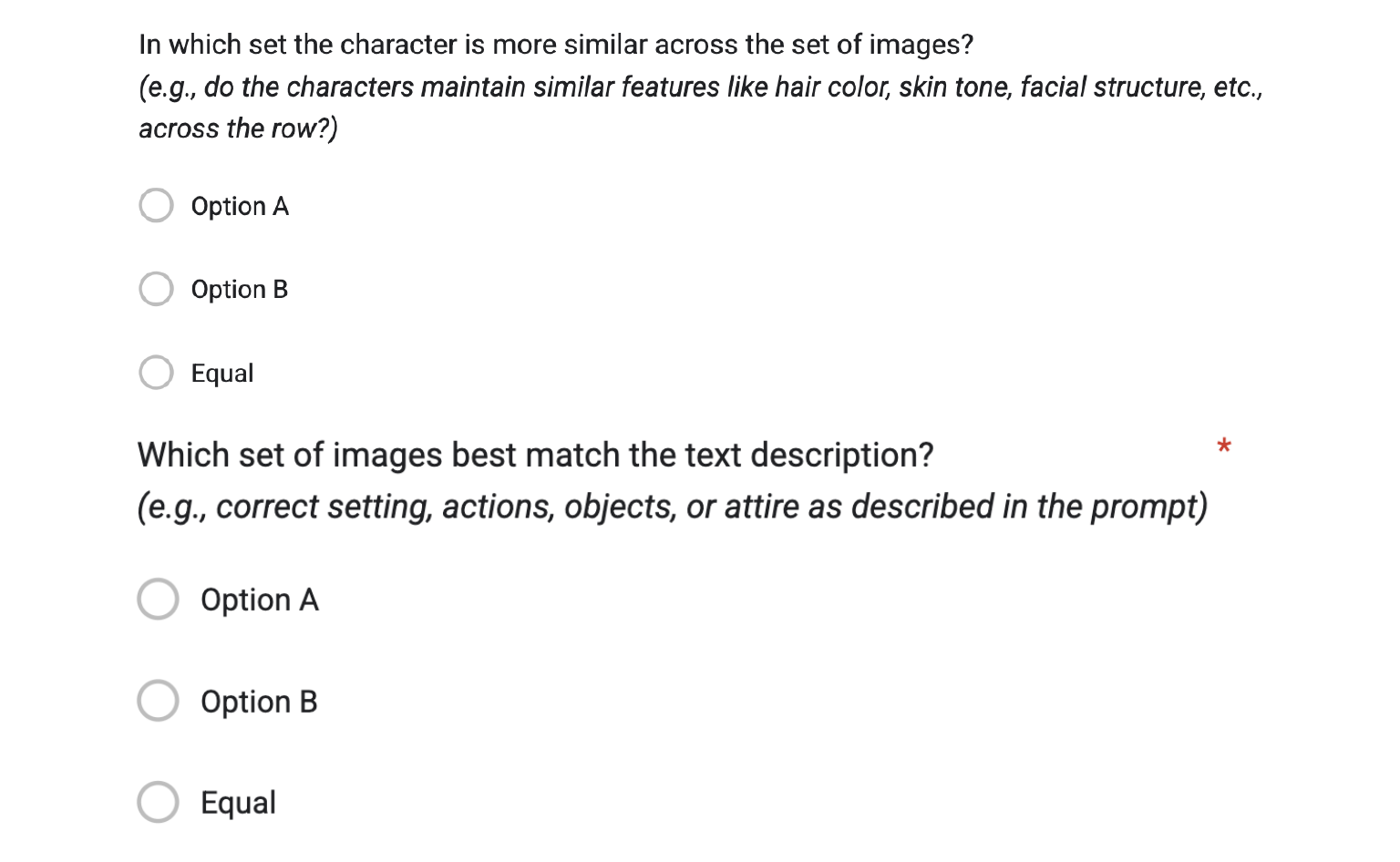}
    \caption{For each set of images, the following evaluation questions were asked in the user study.}

    \label{fig:user_study_questions}
\end{figure*}

Each participant was presented with 12 comparison sets, each containing four randomly selected images per method, as shown in Fig. \ref{fig:user_study_example}. For each set, participants answered three questions corresponding to the three evaluation tasks. An example of the user study questions is shown in Fig.~\ref{fig:user_study_questions}.
\begin{figure*}[hb]
    \centering
    \includegraphics[width=0.67\textwidth]{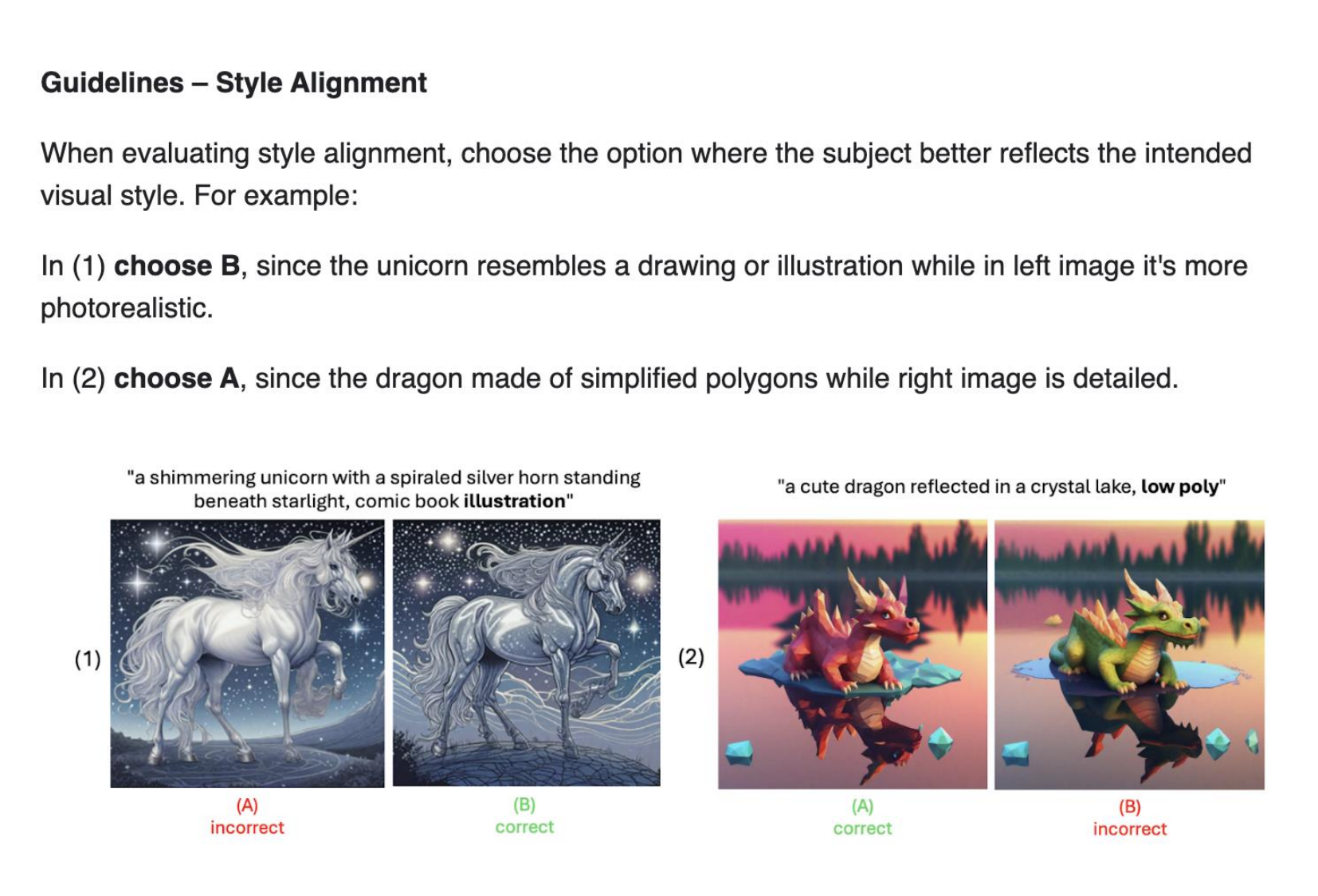}
    
    \centering
    \includegraphics[width=0.67\textwidth]{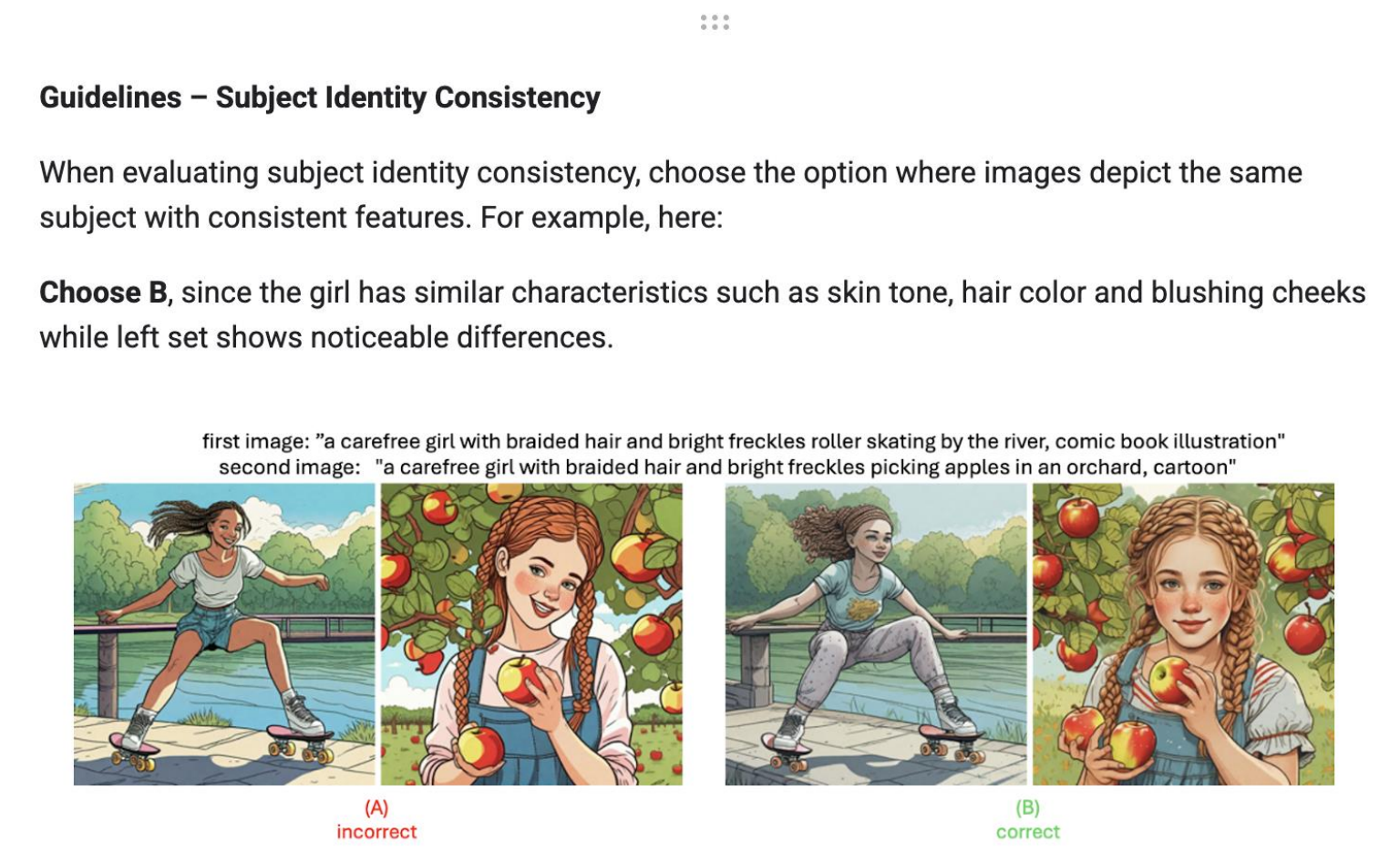}
    
    \label{fig:user_study_struct}
\end{figure*}

\end{document}